\pdfoutput=1
\documentclass[11pt]{article}

\usepackage[preprint]{acl}

\usepackage{hyperref}

\usepackage{times}
\usepackage{latexsym}

\usepackage[T1]{fontenc}
\usepackage[utf8]{inputenc}

\usepackage{microtype}

\usepackage{svg}
\usepackage{inconsolata}

\usepackage{graphicx}
\usepackage{hyperref}
\usepackage{booktabs}
\usepackage{multirow}
\usepackage{array}
\usepackage{amsmath}
\usepackage{scalerel,xparse}

\title{A Retrieval-Based Approach to Medical Procedure Matching in Romanian}
\author{Andrei Niculae\textsuperscript{1}, Adrian Cosma\textsuperscript{1,2}, Emilian Radoi\textsuperscript{1} \\
  \textsuperscript{1}National University of Science and Technology POLITEHNICA Bucharest \\
  \textsuperscript{2}Dalle Molle Institute for Artificial Intelligence Research (IDSIA) \\
  \texttt{\normalsize andrei.niculae1004@stud.acs.upb.ro, adrian.cosma@supsi.ch, emilian.radoi@upb.ro} \\
}

\begin{document}
\maketitle
\begin{abstract}
Accurately mapping medical procedure names from healthcare providers to standardized terminology used by insurance companies is a crucial yet complex task. Inconsistencies in naming conventions lead to missclasified procedures, causing administrative inefficiencies and insurance claim problems in private healthcare settings. Many companies still use human resources for manual mapping, while there is a clear opportunity for automation. This paper proposes a retrieval-based architecture leveraging sentence embeddings for medical name matching in the Romanian healthcare system. This challenge is significantly more difficult in underrepresented languages such as Romanian, where existing pretrained language models lack domain-specific adaptation to medical text. We evaluate multiple embedding models, including Romanian, multilingual, and medical-domain-specific representations, to identify the most effective solution for this task. Our findings contribute to the broader field of medical NLP for low-resource languages such as Romanian.
\end{abstract}

\section{Introduction}
Ensuring accurate mapping between medical procedure names used by different healthcare providers and a standardized terminology set maintained by health insurance companies is a challenging task, with real-world applications. Discrepancies in naming conventions can lead to administrative inefficiencies, misclassification of procedures, and potential barriers for patients seeking insurance coverage. These mismatches can result in denied claims, increased processing times, and overall inefficiencies in the healthcare reimbursement process. For example, "The State of Claims: 2024" report \footnote{\href{https://www.experian.com/healthcare/resources-insights/thought-leadership/white-papers-insights/state-claims-report}{The State of Claims: 2024, Accessed 19.03.2025}} reveals that 46\% of denied claims are due to missing or innacurate data and coding errors. 

\begin{figure}[t!]
    % Link: https://drive.google.com/file/d/15TE-DXieF8yplRzHsOD9jJ-5RcZ9hx4k/view?usp=sharing 
    \centering
    \includesvg[width=\linewidth]{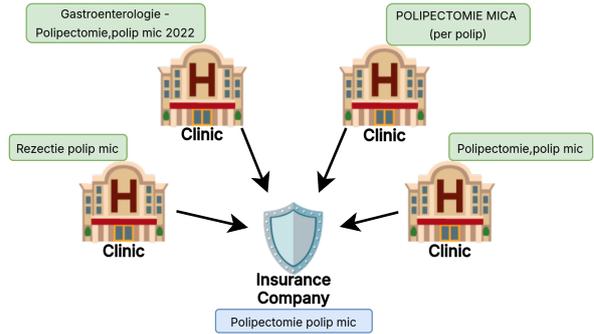}
    \caption{Diagram of the medical procedure matching problem. Clinics often have their own local names for medical procedures that are changed annually, for which a central insurance agency must match to a standardized list of procedures for reimbursement.}
    \label{fig:medmatch-diagram}
\end{figure}

Matching procedure names is similar to the well-known problems of entity resolution and text matching, yet it presents unique challenges in the medical domain. The complexity stems from several factors: \textit{(i)} medical terminology is highly domain-specific and varies across institutions, \textit{(ii)} data distributions are often imbalanced due to the frequency of common procedures overshadowing rare ones, \textit{(iii)} nomenclatures evolve over time, necessitating adaptive matching techniques, and \textit{(iv)} the presence of noise in text data, including typographical errors and abbreviations further complicates standardization efforts. Figure \ref{fig:medmatch-diagram} illustrates this problem. While previous studies have addressed similar challenges \cite{tavabiSystematicEvaluationCommon2024,levyComparisonMachineLearningAlgorithms2022,zaidatCanNovelNatural2024}, most focus on healthcare systems in the United States or other widely studied regions \cite{alexanderOverviewInpatientCoding2003}. International standards are typically adapted by each country, and private insurance companies may develop their own coding schemes, making a universal solution impractical.

This issue is particularly pressing for underrepresented languages such as Romanian. Despite growing interest in NLP for low-resource languages \cite{nigatu-etal-2024-zenos}, Romanian remains significantly underrepresented in medical NLP research. Existing language models such as RoBERT \cite{masala-etal-2020-robert} and RoLLaMA \cite{masala2024vorbecstiromanecsterecipetrain} provide general-purpose Romanian embeddings, but they lack the necessary specialization for medical text processing. Often, for real-world scenarios, multilingual models are ubiquitously used \cite{wangMultilingualE5Text2024,wang2020minilm}, even if they might fail to capture language-specific nuances.

In this paper, we propose a retrieval-based architecture for medical procedure matching. By leveraging metric learning and dense vector representations of procedure names \cite{ramesh-kashyap-etal-2024-comprehensive}, our method can handle a variable number of input-output mappings, can be expanded without retraining the entire model, and integrate efficiently with scalable vector search frameworks such as Milvus \cite{wangMilvusPurposeBuiltVector2021}. This makes retrieval an attractive paradigm for medical name matching, as it enables continuous updates and adaptation to changing medical taxonomies without extensive human intervention. We empirically evaluate three sentence embedding models \cite{wangMultilingualE5Text2024,masala-etal-2020-robert,alsentzer2019clincalbert}, comparing their effectiveness in Romanian medical name matching. 

By focusing on the Romanian healthcare system, our study highlights the broader challenges of medical terminology standardization and provides insights that can inform similar efforts in other low-resource languages. We aim to contribute to the development of robust, scalable, and language-aware retrieval methods for healthcare applications, ultimately improving the efficiency and accessibility of medical insurance systems.

Our contributions are as follows:

\begin{enumerate}
    \item We propose a retrieval-based architecture for matching medical procedure names across different healthcare providers and insurance companies, addressing a pressing real-world problem in the Romanian healthcare system.
    \item We conduct an extensive evaluation of various sentence embedding models, both Romanian \cite{masala-etal-2020-robert}, multilingual \cite{wangMultilingualE5Text2024} and specialized in the medical domain \cite{alsentzer2019clincalbert}, highlighting their performance in the context of Romanian medical text matching.
\end{enumerate}

\section{Related Work}
\noindent \textbf{Sentence embedding models.} 
Semantic text embedding models \cite{ramesh-kashyap-etal-2024-comprehensive} are a significant component of many NLP applications, most notably text retrieval and question answering. Text embeddings are used to capture semantic representations of text that go beyond surface level word and character matching methods such as TF-IDF. Currently, practitioners are using pretrained transformer models such as BERT \cite{reimersSentenceBERTSentenceEmbeddings2019}, either by aggregating word-level representations with a pooling operation, or by using specialized training for text similarity \cite{khattabColBERTEfficientEffective2020}. Currently, the best performing models are aggregated in the MTEB leaderboard \cite{muennighoff-etal-2023-mteb}, a benchmark of several text embedding tasks, including several non-English datasets. For the medical and scientific domain \cite{lewis2020pretrained}, several models have been developed. Models such as SciBERT \cite{beltagy-etal-2019-scibert}, BioBERT \cite{alsentzer2019clincalbert}, ClinicalBERT \cite{alsentzer2019clincalbert} and MedBERT \cite{rasmy2021med} offer domain-specific embeddings by training on either specialized biomedical corpora or task-specific datasets.

However, most contextualized text representation models for the medical domain are focused on the English language, with under-represented languages severely lacking in resources such as specialized models or training datasets. In our setup, medical procedure names are written in Romanian, a low resource language, with only a few pretrained language models \cite{masala2024vorbecstiromanecsterecipetrain,masala-etal-2020-robert}. Currently, for Romanian, only a pretrained RoBERT model \cite{masala-etal-2020-robert} is available for direct contextualized text representations, but no such variant exists for the medical domain. Currently, multilingual models such as E5 \cite{wangMultilingualE5Text2024} and MiniLM \cite{wang2020minilm} are ubiquitously used for non-English tasks.

\noindent \textbf{Medical Procedure Matching.} 
The task of medical procedure matching has been performed in the context of assigning medical notes or pathology reports to a predefined set of medical procedures \cite{tavabiSystematicEvaluationCommon2024,levyComparisonMachineLearningAlgorithms2022,zaidatCanNovelNatural2024}, with a focus on the US medical system.

\begin{table*}[hbt!]
    \centering
    \resizebox{0.75\linewidth}{!}{
    \begin{tabular}{p{0.3\textwidth}|p{0.6\textwidth}}
        \textbf{Masterlist Entry} & \textbf{Associated Clinic Procedures Names}                                                                                  \\
        \toprule
        \multirow{3}{=}{Polipectomie polip mic \\\textit{(Small polyp polypectomy)}}
                                  & {Polipectomie,polip mic \textit{(Polypectomy, small polyp)}}        \\
        \cmidrule{2-2}
                                  % & POLIPECTOMIE MICA (per polip) \textit{(SMALL POLYPECTOMY (per polyp))}                                      \\
        % \cline{2-2}
                                  & Gastroenterologie - Polipectomie,polip mic 2022 \textit{(Gastroenterology - Polypectomy, small polyp 2022)} \\
        \cmidrule{2-2}
                                  & Rezectie polip mic \textit{(Small polyp resection)}                                                         \\
        \midrule
        \multirow{3}{=}{Radiografie omoplat 1 incidenta \\\textit{(Scapula X-ray 1 view)}}
                                  % & Radiografie omoplat fata \textit{(Frontal scapula X-ray)}                                                   \\
        % \cline{2-2}
                                  & Radiografie omoplat (fata sau profil) \textit{(Scapula X-ray (frontal or lateral))}                         \\
        \cmidrule{2-2}
                                  & Omoplat profil \textit{(Lateral scapula view)}                                                              \\
        \cmidrule{2-2}
                                  & RX omoplat profil \textit{(Lateral scapula X-ray)}                                                          \\
        % \cline{2-2}
                                  % & Omoplat fata \textit{(Frontal scapula view)}                                                                \\
        % \cline{2-2}
                                  % & RX omoplat fata \textit{(Frontal scapula X-ray)}                                                            \\
        \midrule
        \multirow{3}{=}{Vitamina B12 \\\textit{(Vitamin B12)}}
                                  % & Vitamina B12 (CLVIT) \textit{(Vitamin B12 (CLVIT))}                                                         \\
        % \cline{2-2}
                                  % & Vitamina B12 (mdcl) \textit{(Vitamin B12 (mdcl))}                                                           \\
        % \cline{2-2}
                                  % & Vitamina B12 (VITB12) \textit{(Vitamin B12 (VITB12))}                                                       \\
        % \cline{2-2}
                                  % & Vitamina B12 (ciancobalamina) - Bioclinica \textit{(Vitamin B12 (cyanocobalamin) - Bioclinica)}             \\
        % \cline{2-2}
                                  & Vitamina B12 serica (5 zile) \textit{(Serum Vitamin B12 (5 days))}                                          \\
        \cmidrule{2-2}
                                  & Vitamina B12 (Cianocobalamina) \textit{(Vitamin B12 (Cyanocobalamin))}                                      \\
        \cmidrule{2-2}
                                  % & Vitamina b12 \textit{(Vitamin B12)}                                                                         \\
        % \cline{2-2}
                                  % & Vit B12 \textit{(Vit B12)}                                                                                  \\
        % \cline{2-2}
                                  % & Vitamina B12 (Ciancobalamina) \textit{(Vitamin B12 (Cyanocobalamin))}                                       \\
        % \cline{2-2}
                                  & ANALIZA SANGE - Vitamina B12 \textit{(BLOOD ANALYSIS - Vitamin B12)}                                        \\
    \end{tabular}}
    \caption{Selected examples of entries in the masterlist and associated procedure names from clinics. There is significant variation in procedure names, which makes simple text matching inappropriate. We provide English translations for convenience.}
    \label{tab:examples}
\end{table*}

\citet{tavabiSystematicEvaluationCommon2024} investigated the problem of mapping unstructured operative notes to Current Procedural Terminology (CPT) codes. The CPT code set is a system used to describe medical, surgical and diagnostic services, that are used for billing and insurance reimbursement processes in healthcare. The authors apply common NLP techniques to assign 44,002 notes to 100 most prevalent CPT codes, treating this problem as a classification task. Using TF-IDF, Doc2Vec \cite{leDistributedRepresentationsSentences2014a} and Clinical Bio-BERT \cite{alsentzer2019clincalbert} embeddings as input they train a support vector machine classifier, for each embedding type. In their experiments, TF-IDF outperformed both BERT and Doc2Vec.

\citet{levyComparisonMachineLearningAlgorithms2022} used machine-learning models for predicting CPT codes from pathology reports. Their study analyzed 93,039 pathology reports from the Dartmouth-Hitchcock Department of Pathology and Laboratory Medicine, classifying 42 CPT codes. They evaluated the performance of XGBoost and BERT—using both diagnostic text alone and all report subfields. Their findings indicated that while BERT outperformed XGBoost when trained only on diagnostic text, but using all report subfields resulted in XGBoost achieving the best performance. 

\citet{zaidatCanNovelNatural2024} have also explored assigning CPT codes to spine surgery operative notes, using XLNet \cite{yang2019xlnet}, a bidirectional LSTM \cite{hochreiterLongShortTermMemory1997} model. They fine-tune the model to their operative note dataset, containing 922 entries.

Previous studies have evaluated the performance of statistical, machine learning and deep learning models on classification of a large number of samples to a relatively small subset of CPT codes. In contrast, we formulate our problem as a retrieval problem, since our dataset is severely imbalanced, and contains two orders of magnitude more CPT codes (38,814 entries). Furthermore, another advantage of this formulation is that by avoiding a fixed set of classes, the addition of more procedures does not require modifying the architecture or retraining the model. Unique to our work, we are the first to tackle this problem in Romanian, a severely low-resource language in terms of specialized models for the medical domain.

\section{Method}
In this section, we provide an overview of the problem description, our dataset of medical procedure names and we describe the architecture for performing mapping between clinic descriptions and a set of standardized procedure names.

\subsection{Problem Description}

\begin{figure*}[hbt!]
    % Draw.io: https://drive.google.com/file/d/11BKJUj00ysBIuxILRDrVS5__cvuNQKBD/view?usp=sharing
    \centering
    \includesvg[width=0.75\linewidth]{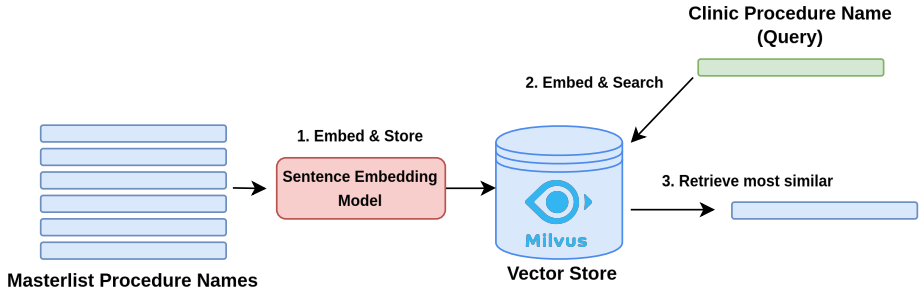}
    \caption{Overall diagram of our method. We formulate medical procedure matching as a retrieval problem: entries in the masterlist are embedded and stored in a vector store and the most similar entry is retrieved based on the similarity with a procedure name from a clinic.}
    \label{fig:inference}
\end{figure*}

The problem of matching medical procedure names to a standardized masterlist is non-trivial. Simple text matching is insufficient, as we will demonstrate in Section \ref{results}. Our dataset is comprised of medical procedures and tests from 528 Romanian private clinics, containing 145,298 unique procedure names mapped to their corresponding masterlist entries. Through manual filtering of incorrect mappings, we reduced the dataset to 139,210 entries. Healthcare providers frequently use varying terms, abbreviations, and phrasing for the same procedure, which creates inconsistencies. To illustrate the difficulty, Table \ref{tab:examples} shows some relevant examples of mappings. Healthcare providers may omit obvious terms, such as "polyp resection" being synonymous with "polypectomy". Similarly, entries such as "frontal or lateral X-ray" must be mapped to "1 view X-ray", as they represent a single test being performed. Terminology variations, like vitamin B12 also being called Cyanocobalamin, can add complexity, especially when descriptions include irrelevant details that mislead text-based matching. Several other relevant examples are presented in Table \ref{tab:bm25-misses}: matching a single medical procedure, when the description actually describes two procedures, not recognizing the semantic meaning of descriptions, ignoring important numerical thresholds, retrieving specific procedures instead of general ones (or vice versa), and prioritizing less important terms.

\begin{figure}
    \centering
    \includesvg[width=\linewidth]{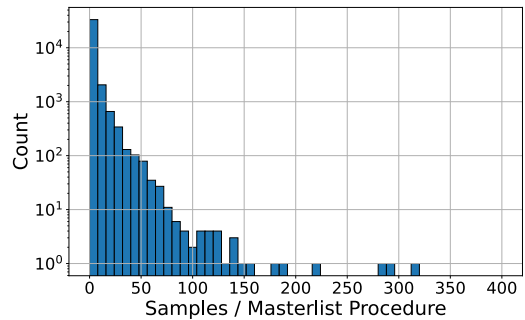}
    \caption{Distribution of number of unique clinic descriptions per masterlist procedure. There is a severe data imbalance: 19,493 (~50\%) out of 39,097 entries contain only a single example.}
    \label{fig:histogram}
\end{figure}

We chose to model our problem as a retrieval problem, and not as a classification problem, since 50\% of elements from the masterlist have only 1 unique procedure assigned.  Figure \ref{fig:histogram} shows the distribution of clinic descriptions assigned to masterlist entries. If we frame our task as a classification problem, we have 39,097 distinct classes, with 19,493 containing only a single sample. Given the severe class imbalance per procedure, a classification model would be inappropriate and would generalize poorly. However, a retrieval-based method can be effectively used by leveraging semantic text embeddings and metric-learning approaches to capture the similarity between clinic descriptions and masterlist entries.

% Greu de explicat asta sa se inteleaga. Am lasat-o simplu ca am filtrat de mana intrarile gresite.
% {\color{orange} During the initial evaluation, we analyze cases where the recommendation model fails to match entries correctly. We find that the model often suggests plausible mappings, whereas the given ground-truth mappings appear to be incorrect. As a result, we fine-tune the embeddings model using the entire dataset and eliminate the misses@100, reducing the dataset to 139,210 entries.}

\subsection{Procedure Matching as Retrieval}
\label{sec:architecture}

In our retrieval setup, we used the provided masterlist procedure names to build a retrieval index \cite{wangMilvusPurposeBuiltVector2021} and clinic descriptions as the references for the queries. We embed the descriptions using dense \cite{masala-etal-2020-robert,wangMultilingualE5Text2024,alsentzer2019clincalbert} and sparse models \cite{bm25}. At inference time, we embed the query clinic descriptions that require a masterlist description and perform a similarity search. The vector DB returns the top-k most similar results for each of our clinic description. Figure \ref{fig:inference} showcases this approach.

The vector index includes two types of entries: masterlist entries and clinic description $\leftrightarrow$ masterlist pairs. In the first scenario, the similarity score is calculated between the query and the masterlist entries, with the index returning the most similar masterlist entries. In the second scenario, the similarity score is computed between the query and the clinic descriptions stored in the index, and the masterlist entry associated with the most similar clinic description is returned. We build our search and evaluation architecture over Milvus \cite{wangMilvusPurposeBuiltVector2021}, a high-performance vector database.

For our setup, we used three types of text embeddings: \textit{(i)} sparse text embeddings using BM25 \cite{bm25}, \textit{(ii)} dense semantic embeddings with several pretrained transformer models \cite{masala-etal-2020-robert,wangMultilingualE5Text2024,alsentzer2019clincalbert}, both zero-shot and fine-tuned with metric learning, and \textit{(iii)} a hybrid ranking approach using RRF \cite{cormackReciprocalRankFusion2009}. 

\subsection{Sparse Embeddings with BM25}
\label{sec:sparse-embeddings}

For computing the sparse embeddings, we use BM25 \cite{bm25} to identify the most relevant word-level features from the training set descriptions and masterlist entries. Text is preprocessed by removing diacritics, punctuation, and Romanian stopwords, followed by stemming the remaining words. At inference time, we compute the inner product between the query descriptions and the masterlist descriptions, as well as the clinic description pairs.

\begin{figure*}[hbt!]
    % Draw.io: https://drive.google.com/file/d/1nxlDirFDHTUeuvX48gcXN8br562So7qc/view?usp=sharing
    \centering
    \includesvg[width=0.75\linewidth]{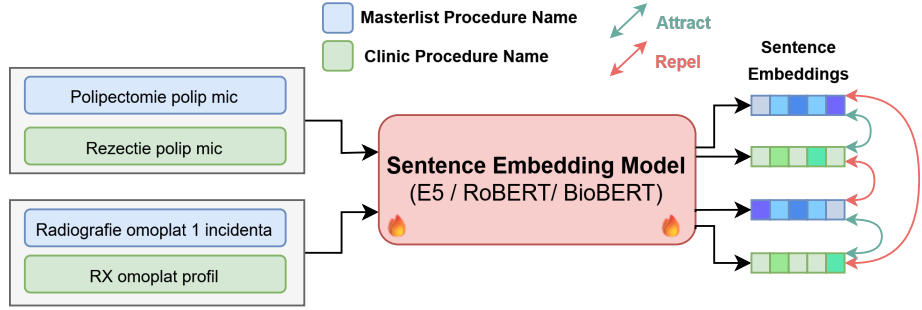}
    \caption{Fine-tuning approach for dense sentence embeddings. A pretrained text embedding model is trained to minimize the distance between representations of masterlist entries and associated clinic procedure names while maximising the distance between every other entry.}
    \label{fig:metric-learning}
\end{figure*}

\subsection{Dense Embeddings with Pretrained Transformer Models}
\label{sec:dense-embeddings-fine-tuning}

Recent studies have highlighted the challenges of selecting optimal sentence embedding models for domain-specific retrieval tasks \cite{wornow2023shaky}. Generic benchmarks do not always align with real-world performance, necessitating task-specific evaluations. The MTEB Leaderboard \citep{muennighoffMTEBMassiveText2023} ranks top-performing embedding models based on retrieval performance across various datasets.

We experiment with three models for dense embeddings: mE5-large \citep{wangMultilingualE5Text2024}, RoBERT-large \citep{masala-etal-2020-robert}, and BioClinicalBERT \citep{alsentzer2019clincalbert}. We select mE5 due to its strong performance on multilingual retrieval tasks, RoBERT as a strong language-specific baseline model pre-trained using only Romanian text, and BioBERT as a domain-specific model, pretrained on biomedical text, which may capture medical terminology better than general-purpose models.

\noindent \textbf{Fine-tuning with Metric Learning.} We fine-tune the pretrained text embedding models using the MultipleNegativesRankingLoss objective \cite{henderson2017efficient}, as shown in Figure \ref{fig:metric-learning}. We consider the clinic descriptions as anchors $(a_i)$ and the corresponding masterlist descriptions $(p_i)$ as positive pairs - $(a_i, p_i)$. The negative pair consists of every combination $(a_i, p_j)$, where $p_j, j \ne i$ are all other masterlist descriptions. In this way, our embedding model learns to increase the cosine similarity between the clinic descriptions and their mapped masterlist description, while decreasing the similarity between the clinic description and all other masterlist items. The model is fine-tuned on 80,911 pairs for 20 epochs, using a batch size of 4096. We use a learning rate of 2e-5, with a cosine scheduler and a warmup ratio of 0.1. All experiments are run on an NVIDIA A100 80GB GPU.

\subsection{Hybrid search}
\citet{cormackReciprocalRankFusion2009} proposed Reciprocal Rank Fusion (RRF) as a method of aggregating the ranking results of multiple information retrieval systems. It is calculated using the formula:

\begin{equation}
    \text{RRFscore}(d \in D) = \sum_{r \in R} \frac{1}{k + r(d)}
\end{equation}

where $D$ is the set of results to be ranked, $R$ represents the multiple returned rankings of these results, $k$ is a constant, and $r(d)$ is the rank of a result $d$. We combine the results of dense and sparse embeddings using RRF and analyze its effect on retrieval accuracy.

\section{Experiments and Results}\label{results}
To evaluate our approach, we split the dataset into a training and evaluation split, containing 80,911 and 58,299 clinic description $\leftrightarrow$ masterlist pairs, respectively. For fine-tuning, we used only the training split. For evaluation, we split the evaluation set into gallery and probe sets in a 4:1 ratio, in a setup similar to 5-fold cross-validation, where gallery entries form the vector store data. Each fold is stratified based on the masterlist entries, such that each fold contains approximately the same distribution of masterlist entries. Specifically, for each masterlist entry, we distribute its associated clinic descriptions evenly across all folds -- for example if 5 clinic descriptions map to the same masterlist entry, each fold will contain exactly 1 such mapping.

\begin{table}[t]
    \centering
    \resizebox{1.0\linewidth}{!}{
    \begin{tabular}{p{3cm}|p{3cm}|p{3cm}|p{3cm}}
        \textbf{\Large Clinic Description} & \textbf{\Large BM25 Miss} & \textbf{\Large mE5 Hit} & \textbf{\Large Observations} \\
        \toprule
        Aplicare sterilet + EEV control & Aplicare sterilet & Montare sterilet (DIU) + ecografie control & Did not account for the additional ultrasound control term \\
        (\textit{IUD application + EEV control}) & (\textit{IUD application}) & (\textit{IUD insertion (DIU) + ultrasound control}) &  \\
        \midrule
        EXOSTOZA & Sonoterapie in exostoze calcaneene & Excizia exostozei & Focused on "exostoses" but did not recognize "excision" as a relevant treatment \\
        (\textit{Exostosis}) & (\textit{Sonotherapy in calcaneal exostoses}) & (\textit{Excision of exostosis}) &  \\
        \midrule
        Chiuretare molluscum contagiosm > 10 leziuni & Chiuretare < 10 leziuni & Chiuretare moluscum contagiosum peste 10 leziuni & Matched on "curettage" but ignored the numerical threshold \\
        (\textit{Curettage of molluscum contagiosum > 10 lesions}) & (\textit{Curettage of < 10 lesions}) & (\textit{Curettage of molluscum contagiosum over 10 lesions}) &  \\
        \midrule
        Radiofrecventa ablatie tumori & Ablatie laser / radiofrecventa tumora ureche dificultate redusa & Excizie leziune cu radiofrecventa & Retrieved a more specific procedure (ear tumor) instead of a general one \\
        (\textit{Radiofrequency ablation of tumors}) & (\textit{Laser/radiofrequency ablation of ear tumor - low difficulty}) & (\textit{Excision of lesion with radiofrequency}) &  \\
        % \midrule
        % Extractie chist sebaceu cu sutura & Extractie chist sebaceu fara sutura & Excizie chirurgicala chist sebaceu+sutura & Matched on "extraction" but did not account for the presence of "suture" (with or without) \\
        % (\textit{Sebaceous cyst extraction with suture}) & (\textit{Sebaceous cyst extraction without suture}) & (\textit{Surgical excision of sebaceous cyst + suture}) &  \\
        % \midrule
        % RMN articulatiisacroiliace & RMN Uro RMN nativ, 3T & RMN articulatii sacroiliace nativ, 1.5T & Retrieved an unrelated MRI modality instead of a joint-specific scan \\
        % (\textit{MRI of sacroiliac joints}) & (\textit{MRI Uro native, 3T}) & (\textit{MRI of sacroiliac joints native, 1.5T}) &  \\
        \midrule
        RM articulatii sacro iliace cu subst. de contrast & Artrodeza articulatiei sacro iliace percutanata cu implant I Fuse & RMN articulatii sacroiliace cu SC, 1.5T & Retrieved a surgical procedure instead of an imaging scan \\
        (\textit{MRI of sacroiliac joints with contrast}) & (\textit{Percutaneous sacroiliac joint arthrodesis with I Fuse implant}) & (\textit{MRI of sacroiliac joints with SC, 1.5T}) &  \\
    \end{tabular}}
    \caption{Selected examples of clinic Descriptions with BM25 Misses, mE5 Dense Embedding Hits. Sparse indexes are not appropriate for this task, which require high level semantic understanding of descriptions.}
    \label{tab:bm25-misses}
\end{table}

\begin{table*}[hbt!]
    \centering
    \resizebox{0.9\linewidth}{!}{
    \begin{tabular}{cc|cccc}
        \textbf{Vector Store Data} & \textbf{Index Type} & \textbf{Acc@1} & \textbf{Acc@3} & \textbf{Acc@5} & \textbf{Acc@100} \\
        \toprule
        \multirow{3}{*}{Masterlist Entries Only} 
        & sparse (BM25) & 52.6 $\pm$ 0.002 & 64.5 $\pm$ 0.002 & 68.5 $\pm$ 0.002 & 86.3 $\pm$ 0.001 \\
        & dense (mE5)    & 78.8 $\pm$ 0.002 & 92.2 $\pm$ 0.002 & 95.0 $\pm$ 0.002 & 99.5 $\pm$ 0.001 \\
        & hybrid (RRF)  & 63.9 $\pm$ 0.003 & 77.7 $\pm$ 0.003 & 82.1 $\pm$ 0.003 & 99.5 $\pm$ 0.001 \\
        \midrule
        \multirow{3}{*}{Masterlist Entries + Mappings} 
        & sparse (BM25) & 68.0 $\pm$ 0.003 & 82.3 $\pm$ 0.001 & 86.1 $\pm$ 0.001 & 94.7 $\pm$ 0.001 \\
        &  dense (mE5)   & \textbf{85.2 $\pm$ 0.003} & \textbf{95.8 $\pm$ 0.001} & \textbf{97.5 $\pm$ 0.001} & \textbf{99.5 $\pm$ 0.001} \\
        &  hybrid (RRF) & 81.0 $\pm$ 0.002 & 92.3 $\pm$ 0.001 & 94.9 $\pm$ 0.001 & 99.5 $\pm$ 0.000 \\
    \end{tabular}}
    \caption{Comparison between sparse embeddings from BM25, dense embeddings from mE5 \cite{wangMultilingualE5Text2024}, and hybrid search, having only masterlist entries in the vector store and having both masterlist and associated clinical mappings. Using dense embeddings from mE5 provides the best results in both cases. Results are averaged across 5 folds.}
    \label{tab:hybrid}
\end{table*}
\begin{table*}[hbt!]
    \centering
    \resizebox{0.9\linewidth}{!}{
    \begin{tabular}{lc|cccc}
        \textbf{Model Name} & \textbf{Type} & \textbf{Acc@1} & \textbf{Acc@3} & \textbf{Acc@5} & \textbf{Acc@100}\\
        \toprule
        RoBERT \cite{masala-etal-2020-robert} & \multirow{3}{*}{off-the-shelf} & 44.7 $\pm$ 0.003 & 53.4 $\pm$ 0.003 & 56.9 $\pm$ 0.004 & 75.3 $\pm$ 0.003 \\
        BioClinicalBERT \cite{alsentzer2019clincalbert} &  & 47.7 $\pm$ 0.003 & 56.7 $\pm$ 0.003 & 60.2 $\pm$ 0.002 & 74.9 $\pm$ 0.003\\
        mE5 \cite{wangMultilingualE5Text2024}  &  & 56.8 $\pm$ 0.003 & 69.4 $\pm$ 0.002 & 74.3 $\pm$ 0.002 & 91.3 $\pm$ 0.002\\
        \midrule
        RoBERT \cite{masala-etal-2020-robert} & \multirow{3}{*}{fine-tuned} & 75.9 $\pm$ 0.001 & 89.9 $\pm$ 0.002 & 93.2 $\pm$ 0.000 & 98.9 $\pm$ 0.001\\
        BioClinicalBERT \cite{alsentzer2019clincalbert} &  & 75.7 $\pm$ 0.002 & 89.2 $\pm$ 0.002 & 92.7 $\pm$ 0.002 & 98.9 $\pm$ 0.000\\
        mE5 \cite{wangMultilingualE5Text2024}  &  &\textbf{ 78.8 $\pm$ 0.002} & \textbf{92.2 $\pm$ 0.002} & \textbf{95.0 $\pm$ 0.002} & \textbf{99.5 $\pm$ 0.001}\\
    \end{tabular}
    }
    \caption{Comparison between different types of text embedding models, having entries in the vector store only from the masterlist entries. We obtained the best results using a fine-tuned version of mE5, a general-purpose multi-lingual model. Results are averaged across 5 folds.}
    \label{tab:masterlist}
\end{table*}

\noindent \textbf{Evaluation Metrics.}
Our primary evaluation metric is Accuracy@k, which measures whether a ground-truth masterlist description is in the first k returned results for a query clinic description. Our target is to optimize for Acc@1, but we also include the results for Acc@3, Acc@5 and Acc@100. In a real-life use of such an system will involve suggesting top-3 or top-5 most similar masterlist entries, and Acc@3 and Acc@5 provides insight into the usefulness of our system. We also include Acc@100, as a low value indicates a problem with the chosen search technique, but usually it indicates the presence of incorrect annotations. In all our results, we show the mean and standard deviation across 5 folds.

\begin{table*}[hbt!]
    \centering
    \resizebox{0.9\linewidth}{!}{
    \begin{tabular}{lc|cccc}
        \textbf{Model Name} & \textbf{Type} & \textbf{Acc@1} & \textbf{Acc@3} & \textbf{Acc@5} & \textbf{Acc@100} \\
        \toprule
        RoBERT \cite{masala-etal-2020-robert} & \multirow{3}{*}{off-the-shelf} & 62.5 $\pm$ 0.005 & 76.8 $\pm$ 0.004 & 81.1 $\pm$ 0.005 & 92.0 $\pm$ 0.004\\
        BioClinicalBERT \cite{alsentzer2019clincalbert} &  & 66.7 $\pm$ 0.005 & 80.8 $\pm$ 0.003 & 84.6 $\pm$ 0.002 & 93.4 $\pm$ 0.002\\
        mE5 \cite{wangMultilingualE5Text2024} &  & 67.9 $\pm$ 0.004 & 85.2 $\pm$ 0.002 & 89.6 $\pm$ 0.002 & 98.1 $\pm$ 0.001\\
        \midrule
        RoBERT \cite{masala-etal-2020-robert} & \multirow{3}{*}{finetuned} & 84.4 $\pm$ 0.002 & 94.8 $\pm$ 0.002 & 96.6 $\pm$ 0.001 & 99.0 $\pm$ 0.001\\
        BioClinicalBERT \cite{alsentzer2019clincalbert} &  & 83.8 $\pm$ 0.003 & 94.3 $\pm$ 0.001 & 96.4 $\pm$ 0.001 & 99.0 $\pm$ 0.001\\
        mE5 \cite{wangMultilingualE5Text2024} &  & \textbf{85.2 $\pm$ 0.003} & \textbf{95.8 $\pm$ 0.001} & \textbf{97.5 $\pm$ 0.001} & \textbf{99.5 $\pm$ 0.001}\\
    \end{tabular}
    }
    \caption{Comparison between different types of text embedding models, having entries in the vector store from both the masterlist entries and associated clinical mappings. We obtained the best results using a fine-tuned version of mE5, a general-purpose multi-lingual model. Results are averaged across 5 folds.}
    \label{tab:historical}
\end{table*}

\subsection{Comparison between different types of search indexes}
In Table \ref{tab:hybrid}, we show a comparison between dense, sparse, and hybrid approaches. For dense embeddings, we used a fine-tuned mE5 \cite{wangMultilingualE5Text2024} model. The results show that the fine-tuned dense model consistently outperforms both sparse and hybrid search methods. When searching only masterlist entries, the dense approach achieves 26.2\% higher Acc@1 than the sparse approach. When using both masterlist and associated mappings, the dense approach obtains a 17.2\% Acc@1 margin. The sparse approach also shows poor performance for Acc@100, indicating that a bag-of-word approach is not appropriate for this task, and semantic understanding is needed. Hybrid search fails to outperform dense search as it is limited by the poor performance of sparse search. In Table \ref{tab:bm25-misses} we show selected examples of clinic descriptions where sparse embeddings fail to capture variations in text descriptions.

\subsection{Fine-tuning with metric learning}
In Table \ref{tab:masterlist} we compare the performance of three dense embedding models: mE5-large \citep{wangMultilingualE5Text2024}, RoBERT-large \citep{masala-etal-2020-robert}, and BioClinicalBERT \citep{alsentzer2019clincalbert}. We obtained that mE5 has higher off-the-shelf retrieval accuracy compared to RoBERT and BioClinicalBERT. This advantage stems from mE5's design as a sentence-transformer model specifically trained to evaluate similarity between sentences or descriptions, whereas RoBERT and BioClinicalBERT is adapted for sentence embedding through a pooling operation over token embeddings.

Sparse search initially outperforms both RoBERT and BioClinicalBERT. However, after fine-tuning, all dense embeddings surpass sparse embeddings in performance metrics, with mE5 maintaining its position as the highest-performing model.

Table \ref{tab:historical} illustrates the impact of incorporating both masterlist and associated mappings in search processes. The inclusion of reduces the performance difference between E5 and the other two models. While the relative ranking of models remains consistent, E5 achieves the highest performance with an Acc@1 of 85.2\% and an Acc@5 of 95\%. Notably, Acc@1 metric may under-represent actual performance. Manual inspection of misclassified results reveals many plausible matches. This discrepancy occurs due to the presence of duplicate entries within the masterlist itself—entries with slightly different formulations that reference identical medical procedures. The markedly higher Acc@3 metric, which captures whether the ground-truth result appears within the first three recommendations, supports this observation. Although duplicated masterlist results present a methodological challenge for evaluation, they do not compromise practical application. The real-world accuracy exceeds the reported metrics, as demonstrated in the next section.

\subsection{Doctor evaluation}\label{doctor-evaluation}
\begin{table}[hbt!]
    \centering  
    \resizebox{0.9\linewidth}{!}{
    \begin{tabular}{l|ccccc}
        \textbf{Model Name} & \textbf{Acc@1} & \textbf{Acc@2} & \textbf{Acc@3} \\
        \toprule
        mE5 - All Data & 94.7 & 98.5 & 99.0\\
    \end{tabular}}
    \caption{Real-world evaluation of our system. Doctors manually evaluated 12,836 new entries after mapping them with a fine-tuned version of mE5 on all data.}
    \label{tab:doctor-results}
\end{table}
Our medical procedure mapping system was used to map new unmapped procedures. We evaluated on new procedure descriptions from 10 clinics, comprising 12,836 unique descriptions. After mapping the procedures using a fine-tuned mE5 models trained on all available data, doctors validated each pair to determine if the masterlist assignment was correct. As shown in Table \ref{tab:doctor-results}, the model achieves a real-world Acc@1 of 94.7\%. The 98.5\% Acc@2 indicates that doctors considered either the first or second recommendation correct, while for only 1\% of entries, doctors assigned a different description than the ones recommended.

Another notable aspect is the speed of the mappings process. While manually mapping the 12,836 descriptions would take more than 60 hours, using our retrieval system reduces this to only 3 minutes, resulting in an 1200$\times$ speedup.

\section{Conclusion}
This paper presents a retrieval-based approach for medical procedure matching in the Romanian healthcare system, addressing the challenges posed by inconsistent naming conventions across clinics and insurance providers. We demonstrate that dense sentence embeddings, particularly fine-tuned multilingual models, significantly outperform traditional sparse methods such as BM25. Our experiments show that a fine-tuned mE5 model achieves the highest retrieval accuracy, with an Acc@1 of 85.2\% when using both masterlist entries and clinical mappings. The real-world evaluation further confirms the efficacy of our approach, achieving a validated accuracy of 94.7\% in a doctor-reviewed dataset. Furthermore, our systems enables significant labor efficiency: using our automated matching systems results in 1200$\times$ speedup compared to manual matching. Our findings contribute to the broader domain of medical NLP for low-resource languages and offer a viable solution for improving the Romanian healthcare system. 

\section*{Limitations}
Our approach has several limitations. Firstly, errors in historical mappings may propagate into future predictions, potentially reinforcing inaccuracies over time. This challenge necessitates periodic human review and correction to prevent systematic errors. Secondly, cosine similarity between embeddings may not always provide a reliable confidence estimate, due to the considerable overlap between the score distributions of hits and misses. This makes it difficult to differentiate between correct and incorrect matches. Incorporating additional uncertainty modeling or ranking refinements could improve result interpretability. Thirdly, while our retrieval model significantly improves over rule-based methods, its performance is still constrained by the lack of a specialized Romanian medical language model. A dedicated medical NLP model trained on domain-specific Romanian corpora could further enhance accuracy.

\section*{Acknowledgement}
This research was supported by the project "Search Techniques for Matching Medical Information - MATCHMED", ID 420246344, and by the project "Romanian Hub for Artificial Intelligence - HRIA", Smart Growth, Digitization and Financial Instruments Program, MySMIS no. 334906. 

\bibliography{refs}

\end{document}